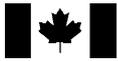

# Self-Replicating Strands that Self-Assemble into User-Specified Meshes

Robert Ewaschuk, National Research Council Canada

Peter D. Turney, National Research Council Canada

February 22, 2005

# *Self-Replicating Strands that Self-Assemble into User-Specified Meshes*

# *Self-Replicating Strands that Self-Assemble into User-Specified Meshes*

## Abstract

It has been argued that a central objective of nanotechnology is to make products inexpensively, and that self-replication is an effective approach to very low-cost manufacturing. The research presented here is intended to be a step towards this vision. In previous work (JohnnyVon 1.0), we simulated machines that bonded together to form self-replicating strands. There were two types of machines (called types 0 and 1), which enabled strands to encode arbitrary bit strings. However, the information encoded in the strands had no functional role in the simulation. The information was replicated without being interpreted, which was a significant limitation for potential manufacturing applications. In the current work (JohnnyVon 2.0), the information in a strand is interpreted as instructions for assembling a polygonal mesh. There are now four types of machines and the information encoded in a strand determines how it folds. A strand may be in an unfolded state, in which the bonds are straight (although they flex slightly due to virtual forces acting on the machines), or in a folded state, in which the bond angles depend on the types of machines. By choosing the sequence of machine types in a strand, the user can specify a variety of polygonal shapes. A simulation typically begins with an initial unfolded seed strand in a soup of unbonded machines. The seed strand replicates by bonding with free machines in the soup. The child strands fold into the encoded polygonal shape, and then the polygons drift together and bond to form a mesh. We demonstrate that a variety of polygonal meshes can be manufactured in the simulation, by simply changing the sequence of machine types in the seed.

## 1  Introduction

Researchers have argued that one of the main objectives of nanotechnology is to manufacture products inexpensively, and that this goal can be effectively achieved by self-replication [2], [8], [9]. We believe that it will be useful to develop computational simulations of self-replicating nanotechnology as engineering tools to assist in the design of actual self-replicating machines.

In our earlier work with JohnnyVon 1.0 (named in honour of John von Neumann [20]), we developed a computational simulation of machines that join to form self-replicating strands (i.e., chains of machines linked by flexible bonds) [18]. These machines drifted about in a virtual liquid, simulated as a two-dimensional continuous space with Brownian motion and viscosity. There were two types of machines, which enabled a strand to encode an arbitrary bit string, by designating one type of machine as representing 0 and the other as 1. Although strand replication faithfully preserved the encoded bit strings, the information in the strings played no functional role in JohnnyVon 1.0. In effect, the simulation had genotypes (i.e., genetic code) without phenotypes (i.e., bodies). From the perspective of potential applications in manufacturing, the absence of phenotypes was a major limitation of JohnnyVon 1.0.

This paper introduces JohnnyVon 2.0, which builds on its predecessor by adding phenotypes to the simulation. The design of JohnnyVon 2.0 was inspired by the work of Seeman on building nanometer-scale structures with DNA [14], [15]. In living organisms, replication is based on DNA (the genotype) and the information encoded in DNA is used to build proteins (the major structural material of the phenotype). Seeman has shown that DNA can serve both as a device for self-replication (genotype) and (surprisingly) as a building material for nanoscopic structures and tools (phenotype). By choosing the appropriate sequence of codons, DNA can be programmed to self-assemble into a wide variety of structures, such as cubes, octahedra, one-dimensional strands, two-dimensional meshes, and three-dimensional arrays. Seeman discusses a variety of potential nanotechnological applications for these structures. For example, a three-dimensional DNA array could facilitate x-ray crystallography, by serving as a scaffolding for holding molecular samples in a regular lattice [15].

Seeman views his work as a first step towards nanorobotics [15]. In Seeman's work, the DNA strands self-assemble, but they do not self-replicate. However, recent work by others has produced a DNA strand that can be replicated by PCR (polymerase chain reaction) and can fold into an octahedron [16]. (Both the replication by PCR and the subsequent folding are assisted by the experimenter, who must add various chemicals at the appropriate stages.)

In JohnnyVon 2.0, there are four types of machines, drifting in a two-dimensional continuous space with Brownian motion and viscosity (i.e., a simulated liquid). The machines exert spring-like attractive and repulsive forces on each other, but internally they are finite state machines. The input to each state machine is based on the presence or absence of bonds with neighbouring machines and on the internal states of bonded neighbours. The internal states govern when bonds are formed or broken and the angles at which bonded machines are joined, and thus determine whether a strand will form a straight line or fold into a specific polygonal shape. The machines are (very roughly) analogous to codons in DNA.

Following the hint of Seeman's work, a strand in JohnnyVon 2.0 serves as both a genotype and a phenotype, at different stages in its career [14], [15], [16]. Like living organisms (but unlike von Neumann's strategy [20]), JohnnyVon 2.0 takes a template-based approach to self-replication. A strand begins its career as a genotype. While acting as a genotype, the strand is approximately straight, so that it can provide a good template for replication. Brownian motion and interactions with other machines will cause the strand to bend slightly, because the bonds between the machines are flexible, but the system is designed so that forces will tend to straighten the strand. Later in its career, the strand may become a phenotype. When this happens, the bonding forces change, causing the bonding angles to alter, and the strand folds. This folding is (approximately) analogous to the way that proteins fold. A folded strand acts as a structural element and is no longer capable of replication.

A typical run of a JohnnyVon 2.0 simulation begins with a soup of unbonded machines and an initial unfolded seed strand of bonded machines. Free (unbonded) machines connect to the seed strand, eventually forming a double strand (two parallel strands). When the new strand is complete, the two strands break apart, and thus we have self-replication. A strand will continue to self-replicate until unbonded machines become scarce. When a strand has not encountered an unbonded machine for a relatively long period of time, the strand stops replicating and folds. The shape that it folds into depends on the types of machines in the strand and their sequential ordering. Folded strands drift in the virtual liquid and bond with each other, forming a mesh. The user can specify the shape of the holes in the final mesh by selecting the sequence of machine types that compose the initial seed strand.

In Section 2, we discuss related work with von Neumann's universal constructor, self-replicating loops, and artificial chemistry. We compare JohnnyVon 2.0 to the related work, in terms of the goal of assisting with research and development of applications in nanotechnological manufacturing. The details of JohnnyVon 2.0's design are explained in Section 3, including the changes that have been made from JohnnyVon 1.0 [18]. We present our experiments in Section 4. Each experiment is a run of the simulation with an initial seed strand. We demonstrate that a variety of polygonal meshes can be manufactured by varying the initial seed. Section 5 examines limitations of JohnnyVon 2.0 and problems and projects for future work. Potential applications are suggested in Section 6 and we conclude in Section 7.

## 2   Related Work

Sipper provides a good survey of research on self-replication [17]. Some of the research involves actual mechanical devices and some is based on organic chemistry, but we restrict our discussion here to computer simulations of self-replication. We compare JohnnyVon 2.0 to von Neumann's universal constructor [10], [20], self-replicating loops [6], [11], [12], [13], [19], and artificial chemistry [4], [5]. We consider the degree to which each is useful as an engineering tool to assist in the design of actual (physical) self-replicating machines.

### 2.1   Universal Constructor

Von Neumann's approach to self-replication was to design a universal constructor, which could build anything, and therefore could build itself as a special case [20]. He described five different models (i.e., five different kinds of simulations), with varying levels of realism and concreteness. The design of the universal constructor was only worked out in detail for the cellular automata model, which was the most abstract of the five models.

The cellular automata model consisted of a two-dimensional space divided into a grid of discrete cells. Each cell was a finite automaton with twenty-nine possible states. A run of the simulation begins by setting the initial states of the cells, and then the states change in a sequence of discrete time steps, according to a fixed set of state transition rules. Transitions between states for a cell depend on the states of its neighbouring cells.

In this model, the universal constructor was composed of a group of several thousand cells that begin in a specific configuration of initial states. Another line of cells acts as a kind of tape, which is read by the universal constructor. For any given finite configuration of cell states, there is a tape that can cause the universal constructor to build the given configuration. As a special case, there is a tape that can cause the universal constructor to build a copy of itself, thereby self-replicating.

Although the computers available to von Neumann were not sufficiently powerful to run this model, it has since been successfully implemented [10]. However, even on a modern computer, it would take many weeks of continuous operation for a universal constructor to make a single copy of itself and its tape.

### 2.2   Self-Replicating Loops

Langton demonstrated self-replication in a cellular automata model that was much simpler than von Neumann's model [6]. He achieved this simplification by designing a constructor that could construct only itself, instead of trying to make a universal constructor. His cellular automata model had eight states instead of twenty-nine and his constructor was composed of a group of about a hundred cells in a specific initial configuration, instead of a group of several thousand cells.

In Langton's model, the cells of the constructor are arranged in a loop. The states of the cells in the loop go through a cycle, periodically creating a copy of the original loop. Starting from the initial loop, increasing numbers of copies spread across the grid.

The idea of self-replicating loops in cellular automata models has since been developed further by many researchers [11], [12], [13], [19]. Self-replicating loops have exhibited many interesting behaviours, including evolution [11], [13] and self-repair [19].

### 2.3    Artificial Chemistry

Hutton introduced self-replication in an artificial chemistry simulation, using a template-based approach [4]. A chain of molecules forms a template against which other molecules bond, similar in concept to JohnnyVon 1.0 [18]. A run of the simulation begins with a seed chain in a soup of free molecules. By a series of chemical reactions, a parallel chain of molecules forms next to the seed chain. When the parallel chain is complete, it separates from the seed chain and process repeats.

Hutton's first approach was a cellular automata model [4], but the discrete space constrained the mobility of the simulated molecules, hence Hutton's second approach used a continuous space [5], like JohnnyVon 1.0 [18]. In Hutton's second model, molecules move in a continuous two-dimensional space, following linear trajectories until an obstacle (e.g., the container wall or another molecule) is encountered (i.e., the motion is a billiard ball model). When molecules make contact with each other, they undergo a chemical reaction that bonds them together, according to the rules of the artificial chemistry.

In Hutton's first model [4], the molecules only replicate, but in his second model [5], they also build a circular barrier, suggestive of a cell wall. Each time a chain replicates, the new chain builds a wall around itself.

### 2.4    Suitability for Nanotechnological Manufacturing

Given our objective, to contribute to nanotechnological manufacturing, there are four factors that are particularly relevant for discussing related work in computational simulations of self-replication:

**Realism:** the degree of physical realism of the simulation (*low*, *medium*, or *high*), which determines the efficacy of the simulation as a tool for engineering of nanotechnology;

**Genotype/Phenotype:** the presence or absence of a distinction between genotype and phenotype (*yes* or *no*), which determines the flexibility of the system for applications in manufacturing;

**Programmability:** the degree of programmability by the user (*low*, *medium*, or *high*), without altering the basic rules of the simulation, which also affects the flexibility for manufacturing purposes;

**Tractability:** the degree of computational tractability of the simulation (*low*, *medium*, or *high*), which determines the feasibility of running interesting simulations on current computer hardware.

Using these four factors, we compare JohnnyVon 2.0 to related work in Table 1. This table is intended to informally summarize the points that we discuss in the following paragraphs.

Table 1. Comparison with related work.

| System | Realism | Genotype/Phenotype | Programmability | Tractability |
| --- | --- | --- | --- | --- |
| Universal constructor | low | yes | high | low |
| Self-replicating loops | low | no | low | high |
| Artificial chemistry | medium | yes | low | high |
| JohnnyVon 1.0 | medium | no | low | high |
| JohnnyVon 2.0 | medium | yes | medium | high |

Von Neumann's approach makes a strong distinction between the phenotype (the group of cells that compose the universal constructor) and the genotype (the line of cells that compose the instruction tape, read by the universal constructor). The universal constructor was designed to be universal; it is programmable to the highest possible degree (ignoring configurations with infinite numbers of cells). However, the cellular automata model is not realistic; its high level of abstraction makes it difficult to use in actual manufacturing applications. Von Neumann recognized this limitation of cellular automata, which is why he felt the need for four other (more concrete) models [20]. The complexity (low tractability) of the universal constructor is also a barrier to applications. Unfortunately, increasing the realism of the models is likely to also increase the complexity of the universal constructor.

Self-replicating loops are much less complex than von Neumann's universal constructor, but this simplicity comes with a loss of programmability. There is no meaningful difference between genotype and phenotype in the loops. Also, self-replicating loops are not usefully user-programmable. If the user makes any changes to the states of the group of cells that compose the initial seed loop, it is likely that the seed will no longer be able to self-replicate, or it will repair itself, undoing the user's change [19]. Additionally, like von Neumann's universal constructor, self-replicating loops have only been implemented in cellular automata models (although it is conceivable that they could be implemented in other models), thus their high level of abstraction limits their practical use in manufacturing applications.

Hutton's most recent work with self-replication in artificial chemistry has both genotype (self-replicating strands of molecules) and phenotype (the circular barrier they build, like a cell wall) [5]. The simulation is also considerably more concrete and realistic than cellular automata models. Hutton's latest model includes virtual chemistry, continuous space, mobile molecules, and a simple virtual physics (idealized billiard ball physics). However, the degree of user-programmability is low. It seems that it is not possible for the user to control what structures are constructed by the system (e.g., circular barriers) by designing the initial seed chain. The structures appear to be hard-coded in the rules of the artificial chemistry.

JohnnyVon 1.0 has moderate physical plausibility, with a virtual physics that includes continuous space, Brownian motion, viscosity, momentum, and attractive and repulsive fields [18]. It can be interpreted as a model of nanobots floating in a thin layer of liquid. However, it has no distinction of genotype and phenotype; the strands can only self-replicate. The user can encode arbitrary information in a seed strand, but this does not constitute programmability, since the encoded information has no functional role in the simulation.

JohnnyVon 2.0 augments its predecessor with a clear distinction between genotype and phenotype. This distinction separates the act of self-replication from the act of building useful structures. The separation of these activities allows the user to program the construction without worrying about the impact of the programming on self-replication. We demonstrate (in Section 4) that the user can program JohnnyVon 2.0 to build a variety of polygonal meshes, by specifying the sequence of machine types in the initial seed strand; the user does not need to make any changes to the rules of the virtual physics. However, the variety of structures that can be built in JohnnyVon 2.0 is still considerably less than what can be done by von Neumann's universal constructor.

## 3   JohnnyVon 2.0

We first give an informal description of the design objectives for JohnnyVon 2.0 and briefly describe the mechanisms used to achieve them. We then cover the few fundamental changes to the model used in the original version of JohnnyVon. In the remainder of this section, we examine the new features of JohnnyVon 2.0 in detail. We encourage the reader who is unfamiliar with JohnnyVon to begin by viewing Figure 1 in Section 4.1. This figure should make it easier to understand the following discussion.

### 3.1   Summary of JohnnyVon 1.0

In the original JohnnyVon, we presented a simulation in which independent objects floated in a virtual two-dimensional liquid [18]. In JohnnyVon 1.0, we called these objects *codons,* but we now prefer to call them *machines*, since their role in JohnnyVon 2.0 has expanded beyond encoding bits. These machines are intended to be an abstract representation of nanobots (simple nanometer-scale robots). The machines can bond together to form *strands*. When a strand is placed in a soup of free (unbonded) machines, the free machines bond to the strand, eventually resulting in two parallel strands, in which the new strand is a mirror image of the original strand. When the new strand is complete, it splits away from the original strand. The new strand then begins replicating along with the original seed strand, until all free machines are bonded in strands. In JohnnyVon 1.0, there were two types of machines, which we called type 0 and type 1. Given two types of machines, the strands can encode and replicate arbitrary binary strings.

### 3.2   Design Objectives for JohnnyVon 2.0

In our outline of future work for JohnnyVon 1.0, we noted that the absence of phenotypes was a significant limitation [18]. Our main design objective for JohnnyVon 2.0 was to add phenotypes. Furthermore, we wanted the phenotypes to be user-programmable. That is, we wanted the user to be able to specify the structure of the phenotypes by encoding instructions in the seed strand, rather than by modifying the rules of the simulation. We believed that these changes would make JohnnyVon 2.0 more suitable for applications in nanotechnological manufacturing.

### 3.3   Overview

In living organisms, genotypes and phenotypes are composed of different materials. Roughly speaking, genotypes are composed of DNA and phenotypes are composed of protein. In JohnnyVon 2.0, following Seeman [14], [15], we decided to use the same material for both purposes. Machines in JohnnyVon 2.0 sometimes behave as genotypes (they participate in self-replication) and sometimes behave as phenotypes (they serve as structural elements), depending on the circumstances.

Using one material for both purposes seemed simpler to us than using two materials. We believe that it also has advantages for manufacturing. If there were two types of materials, the finished product of the manufacturing process would either contain a blend of the two materials or there would need to be a method for removing the unnecessary genotype material. In the first case, where there is a blend, the genotype material is wasted. It consumes resources, adds weight to the final product, and serves no purpose in the final product. In the second case, where the genotype material is removed from the final product, the removal step adds extra complexity to the manufacturing process and consumes extra energy. However, using the same material for both genotypes and phenotypes implies that the finished product contains unnecessary computational power.

JohnnyVon 2.0 has four types of machines, in contrast with two types in JohnnyVon 1.0. When a strand of bonded machines is acting as a genotype, it forms a straight line (ignoring random perturbations from Brownian motion). When acting as a phenotype, it folds into a polygonal shape. In a typical run of the simulation, an initial seed strand is in a soup of free machines. We designed JohnnyVon 2.0 so that the strands tend to self-replicate (they act as genotypes) until free machines become scarce, at which point they tend to fold into structural elements (they act as phenotypes). There is no central controller that tells the strands when to fold; all control is local and distributed. The machines in each strand keep track of how much time has passed since they last bonded to a free machine, and they share this information with their immediate neighbours in the strand. If a strand has failed to self-replicate and no free machines have recently joined the strand (presumably due to a lack of free machines), after a certain amount of time, the strand will release whatever machines are bonded to it (the partially-built mirror strand) and fold up.

The folding angle between any two bonded machines in a strand is governed by their respective types. Since there are four types of machines ($n$), there are sixteen possible pairs of machine types ($n^2$). The order of the types in a pair does not affect the folding angle of the pair. Since there are ten possible unordered pairs ($n(n+1)/2$), there could be ten different folding angles. Given various operational constraints, JohnnyVon 2.0 has five different folding angles.

Once folded up, the polygonal shapes bond together to form a mesh. We designed JohnnyVon 2.0 so that one seed will only form one mesh. This gives the user more control over the manufacturing process. The user can regulate the number of meshes that will be produced in a given batch by regulating the number of seeds. To implement this constraint, the first child strand that is produced by the initial seed strand (the *genotype seed* strand) immediately folds and acts as a seed for the formation of the mesh (i.e., it becomes the *phenotype seed*). When other strands fold later in the simulation, they cannot bond with each other, they can only bond with the phenotype seed. Once they have bonded with the phenotype seed, they change state, so that free phenotype strands (freshly folded, unbonded strands) are now allowed to bond with them. Thus the mesh grows around the first child strand, and one genotype seed strand yields only one final mesh. (A consequence of this implementation is that multiple seeds are also likely to result in only one mesh.)

As we show in Section 4, JohnnyVon 2.0 supports the three regular polygons that can tile the plane: triangles, squares, and hexagons. It also supports rectangles, octagons, and larger versions of each of the polygons, with three or more machines per side.

### 3.4 Basic Changes

This subsection describes the core changes in the simulation that were made from JohnnyVon 1.0 to JohnnyVon 2.0.

### 3.4.1   Change of Research Focus

In our earlier work with JohnnyVon 1.0, we were attempting to build a simulation that would interest (at least) three research communities: biologists interested in the origins of life, nanotechnologists, and artificial life researchers in general. JohnnyVon 1.0 was deliberately designed to allow mutation, since mutation is required for evolution. With mutation, we were able to demonstrate both spontaneous unseeded self-replication (interesting for studying the origins of life) and limited evolution (interesting to artificial life research in general) [18].

In our work with JohnnyVon 2.0, we became aware of a conflict in our goals. In particular, mutation is desirable for living organisms, but it is typically not desirable in a manufacturing process. We decided to focus on nanotechnological manufacturing and give up biological plausibility. Therefore JohnnyVon 2.0 was designed to discourage mutation. In a continuous-space simulation (in contrast to a cellular automata simulation), it is very difficult to eliminate all possible sources of error in self-replication and self-assembly, thus they can still occur in JohnnyVon 2.0, but we have tried to minimize errors.

The decision to focus on manufacturing also made it easier for us to choose a single material for both genotypes and phenotypes, instead of the more biologically plausible two-material approach. However, it is possible that life began with a single material for both purposes (e.g., RNA [7]), so our research may still have some interest for those who study the origins of life.

### 3.4.2   Variable Field Sizes

Bonds between machines are formed by spring-like attractive fields. Part of the mechanism that was in place to support mutation in JohnnyVon 1.0 was a variable field size. In certain circumstances, the field would be small, permitting rare accidental bonds, while other times it would be large, to strengthen intentional bonds. The accidental bonds were a cause of mutation (replication errors), whereas the intentional bonds were part of faithful replication.

In JohnnyVon 2.0, field sizes do not change. Fields attract, repel, or ignore other fields, but they have a constant circle of influence. In situations where the original version has small fields, the new version has inactive fields. This modification substantially reduces the likelihood of mutations.

### 3.4.3   Physical Constants

The physical constants for viscosity, Brownian motion, and motion dampening were changed to suit the new requirements. The values of these constants were experimentally tuned to achieve our design objectives while maximizing the speed (computation efficiency) of the simulation.

### 3.4.4   Arms and the Machine

In JohnnyVon 1.0, each machine was shaped like a capital letter 'T'. Each machine had four arms, but two of the arms overlapped (along the vertical bar of the T), so the figures in the paper seem to show three arms [18]. Each arm had an attractive or repulsive field with a circular shape, centered on the tips of the arms. The fields were colour coded, and we named the arms according to the colours of their associated fields.

In JohnnyVon 2.0, each machine is shaped like a plus sign '+' (see Figure 1 in Section 4.1). Each machine now has five arms, but two of the arms overlap, so the figures seem to show four arms. It is no longer convenient to refer to the arms by the colours of their fields. We now refer to the arms by their relative positions (up, left, right), when the machine is rotated into a canonical position. The lengths of the arms have been modified to facilitate building meshes.

### 3.5　Definitions

The following definitions will help to clarify the discussion.

**Machine:** the basic objects in the JohnnyVon 2.0 simulation. There are four types of machines, numbered 1 through 4. All four types are shaped like a plus sign '+'.

**Arm:** each machine has five arms, but two of the arms overlap, so the figures seem to show four arms. In the figures, the arms are represented by black lines.

**Canonical position:** the machines are mobile and can rotate at any angle, but it is convenient to describe them when they are rotated into a standard reference position, which we call the *canonical position*. In the canonical position, the shortest arm points down, the two longest arms point right and left, and the medium-length arm points up. Another short arm points up in canonical position, but it is hidden by the medium-length arm that points up.

**Left and right arms:** the two longest arms, pointing left and right when the machine is in canonical position. When machines bond to form a strand, adjacent machines in the strand are bonded to each other at the tips of their left and right arms.

**Up arm:** the longer of the two arms that point up when the machine is in canonical position. When a strand replicates by forming a mirror strand, the machines in the mirror strand are bonded to their neighbours in the original strand at the tips of their up arms. Also, when the strands fold into polygons and join to form a mesh, the polygons bond to each other at the tips of their up arms.

**Repellor arm:** the shorter of the two arms that point up when the machine is in canonical position. This arm overlaps the up arm in the figures, so it is not visible. When a strand has completely replicated, repulsive fields are briefly activated at the tips of the repellor arms. This splits the original strand from the mirror strand and pushes the two strands apart.

**Overlap detector arm:** the short arm that points down in canonical position. This arm is used to detect when two folded strands (e.g., polygons) overlap in a mesh.

**Container:** the space that contains the machines. Machines move about in a two-dimensional continuous space, bounded by a grey box. The centers of the machines are confined to the interior of the grey box.

**Liquid:** a virtual liquid that fills the container. The trajectory of a machine is determined by Brownian motion (random drift due to the liquid) and by interaction with other machines and the walls of the container. The liquid has a viscosity that dampens the momentum of the machines.

**Soup:** liquid with machines in it.

**Field:** an attractive or repulsive area associated with a machine. The range of a field is bounded by a circle. In addition to attracting or repelling, a field can also exert a bending force, which twists the machines to form a particular angle. A field's interaction (attract, repel, or ignore) with another field is determined by many factors, including the type and state of each machine. The fields behave somewhat like springs.

**Tip:** the outer end of an arm, where the fields are centered.

**Middle:** the inner ends of the arms, where the five arms meet. This is not the machine's geometrical center, but it is treated as the center of mass in the simulation.

**Bond:** machines can bond together when the field of one machine intersects the field of another. Not all fields can bond. This is described in detail later.

**Up (left, right) neighbour:** the machine that is bonded to the up (left, right) arm of a given machine.

**Strand:** a chain of machines joined by left arm to right arm bonds.

**Unfolded strand:** during replication, the bond angles are such that the replicating strand tends to be straight. Brownian motion and other forces perturb the strand, so it cannot be perfectly straight, but twisting forces in the bonding fields tend to straighten the strand, so it is rarely far from being straight.

**Folded strand:** under specific circumstances (described below), each bond in a strand will change its angle, causing the strand to fold.

**Gene:** another name for an unfolded strand. A strand in its genotype state. The strand will form a straight line (approximately).

**Phene:** another name for a folded strand. A strand in its phenotype state. The strand will form a closed loop.

**Seed gene:** an unfolded strand that is added to a soup of free machines, to initiate the process of self-replication.

**Seed phene:** the first child of the seed gene forms the seed phene, which acts as a starting point for the growth of the mesh. This ensures that one seed phene will yield only one mesh.

**Mesh:** a group of phenes bonded together.

**Free machine:** a machine with no bonds.

**Sideways bond:** a left neighbour to right neighbour bond. Machines in a strand (both phenes and genes) are joined by sideways bonds.

**Up bond:** an up neighbour to up neighbour bond. Phenes in a mesh are joined by up bonds. During replication, a parent gene is joined to its partially constructed child gene by up bonds.

**Time:** the number of steps that have been executed in a run of the simulation, since the initialization of JohnnyVon. The initial configuration is called step 0 (or time 0).

**Tolerance:** each bond has a desired angle (which changes when a strand folds). The two machines that participate in each bond have a tolerance for the difference between the current angle and the desired angle. Forces can push bonds out of tolerance. Bonds that are consistently out of tolerance can break.

**Counter:** a special piece of information stored in each machine that normally increments during each time step. Each machine has several counters.

**State:** the combination of internal information (counters, bonds, and other state variables) and external relationships (position, rotation, and velocity) that determines the behaviour of a machine.

### 3.6 Machine States

The state of a machine is represented by a vector. The vector elements that represent internal aspects of the machine are all discrete. The vector elements that represent external relationships between machines are mostly continuous. The discrete, internal elements are governed by state transition rules that are applied in discrete timesteps. The continuous, external elements are governed by the laws of the virtual physics. The physical laws are inherently continuous, but they are necessarily approximated discretely in any computational simulation.

Internal state information includes various flags, counters, and state variables, as summarized in Table 2. External state information includes spatial location and orientation, angular velocity, linear velocity, the presence or absence of bonds with other machines, and bonding angles, as given in Table 3. Some derived variables are shown in Table 4. The derived variables are calculated from state variables.

Table 2. Variables for elements of the state vector that represent internal aspects of the machine.

| Variable name | Range | Description |
| --- | --- | --- |
| *type* | {1, 2, 3, 4} | • type of machine <br> • static for a given machine |
| *id* | {0, 1, 2, …} | • unique identifier for machine <br> • static for a given machine |
| *fold-counter* | {0, 1, 2, …} | • used to decide when the strand should fold |
| *repel-counter* | {0, 1, 2, …} | • during splitting, controls how long the repellor arms of a strand are active |
| *stress-counter* | {0, 1, 2, …} | • counts the time since a machine was last *in-tolerance* |
| *strand-position* | {1, 2, 3} | • used to decide where a machine is in a replicating strand <br> • described in detail elsewhere [18] |
| *split-state* | {1, 2, 3, 4} | • used to determine when to split <br> • described in detail elsewhere [18], except that *split-state* now has a fourth value, indicating that *shatter* should be set to 1 (true) |
| *reset-counter* | {0, 1} | • indicates that the fold-counter should be reset |
| *fold-now* | {0, 1} | • indicates that each machine in the strand should set its *folded* flag |
| *unfold* | {0, 1} | • indicates that a phene should unfold <br> • this occurs when one phene overlaps another in a mesh |
| *seed-gene* | {0, 1} | • flag for identifying the seed gene <br> • the seed gene never folds, in case more free machines become available for replication |
| *seed-phene* | {0, 1} | • flag for making the seed phene |
| *in-mesh* | {0, 1} | • indicates whether this machine is connected to the mesh |
| *replicated* | {0, 1} | • 1 (true) if and only if the machine has been through a successful replication. <br> • in particular, machines in the seed gene have not replicated at the start of a simulation |
| *shatter* | {0, 1} | • indicates that the machine should break all bonds and return to being a free machine. |
| *folded* | {0, 1} | • if 1, the machine tries to form angular bonds with its left and right neighbours <br> • if 0, the machine tries to form straight bonds with its left and right neighbours |

Table 3. Variables for elements of the state vector that represent external relations.

| Variable name | Range | Description |
|---|---|---|
| *x-position* | $\Re$ | • state of the machine with respect to the container |
| *y-position* | $\Re$ | • vary due to Brownian motion, viscosity, and forces from interactions between fields |
| *angle* | $[0, 2\pi]$ | |
| *x-velocity* | $\Re$ | |
| *y-velocity* | $\Re$ | |
| *angular-velocity* | $[0, 2\pi]$ | |
| *left-neighbour* | {0, 1, 2, …} | • identifier of the machine (if any) bonded to the named arm |
| *right-neighbour* | {0, 1, 2, …} | |
| *up-neighbour* | {0, 1, 2, …} | |

Table 4. Derived variables that are calculated from state variables.

| Variable name | Range | Description |
|---|---|---|
| *in-tolerance* | {0, 1} | • indicates whether each existing bond is within a certain (fixed) tolerance of the desired angle<br>• used to help avoid making bonds when the machine is in a potentially unstable situation. |
| *bend-location* | {1, 2, 3, 4} | • indicates where a machine is located in a phene |

Machines that are directly bonded together can sense each other's states. This is analogous to cells in cellular automata, which can sense the states of their immediate neighbours in the grid. The state transition rules and the virtual physics are local, in the sense that there is no global control structure. No machine can directly sense the state of another machine unless they are directly connected, although state information can be passed neighbour-to-neighbour along a strand. No machine can directly exert a force on another machine unless the circular boundaries of their fields overlap, although forces can be passed neighbour-to-neighbour along a strand.

Most of the state transition rules and physical laws in JohnnyVon 2.0 are carried over from JohnnyVon 1.0 without change. The details of JohnnyVon 1.0 are fully described elsewhere [18]. The changes we made in JohnnyVon 2.0 were outlined above, in Section 3.4.

### 3.7   New Rules

The following subsections describe the rules that are new in JohnnyVon 2.0.

### 3.7.1   Folding

The leftmost machine in an unfolded strand determines when the strand will fold. A machine knows it is leftmost when it has a *right-neighbour* but no *left-neighbour*. As part of its internal state, each machine maintains a *fold-counter*. After a strand has replicated and split, the *fold-counter* in each machine in the newly formed strand starts counting. When a machine gains an *up-neighbour*, it triggers a *reset-counter* signal. This signal is passed to the *left-neighbour,* down the strand, until it reaches the leftmost machine. When the leftmost machine receives the signal, it resets its *fold-counter* to 0. In this way, as long as a replicating strand continues to receive new *up-neighbours*, it will not fold up. If the strand successfully replicates, the leftmost *fold-counter* in each of the two new strands is set to 0.

Once the *fold-counter* in the leftmost machine hits a fixed upper limit, that machine triggers a *fold-now* signal. This signal is passed down the strand, setting the *folded* flag as it goes. This causes the strand to fold up according to the types of each of the two machines involved in a sideways bond.

When a strand folds, there are typically some *up-neighbours* attached to the folding strand, as the folding strand has usually partially replicated itself. Machines with an *up-neighbour,* but having a false (0) *replicated* flag, monitor their *up-neighbour*'s *folded* flag. If this flag becomes true (1), then such a machine will set its own *shatter* flag to true, and thus release all its bonds. (Shattering is described in detail in 3.7.7.)

Machines with the *seed-gene* flag set to 1 never fold, thus the initial seed strand is always available, to continue replicating as soon as there is a supply of free machines of the right types. This is a safeguard against situations in which a temporary scarcity of free machines persists for longer than the fixed upper limit on *fold-counter*.

### 3.7.2 Angles and Bonds

In contrast to JohnnyVon 1.0, we now have four types of machines instead of two types, and the type of a machine affects its behaviour in both genes and phenes. In genes, the types govern bonding during replication, where the rule is simply, "Likes attract, others are ignored." More formally, the bonding rule for up bonds in genes is given in Table 5.

Table 5. Pairs of machine types that will permit an up bond when in genes.

| Bond | 1 | 2 | 3 | 4 |
|---|---|---|---|---|
| 1 | + |   |   |   |
| 2 |   | + |   |   |
| 3 |   |   | + |   |
| 4 |   |   |   | + |

This rule implies that each replicated strand is a mirror of its parent, rather than an exact copy. For example, a strand of types 1-2-3, reading left to right, in canonical position, will replicate as 3-2-1. The phenes that we demonstrate here (in Section 4) have this simple symmetry (i.e., the strands and their mirror images both fold into the same polygonal shapes), so this is not a problem.

The rules for up bonds in genes are different from the rules for up bonds in phenes. In phenes, only certain combinations of types will bond on their respective up arms, as given in Table 6.

Table 6. Pairs of machine types that will permit an up bond when in phenes.

| Bond | 1 | 2 | 3 | 4 |
|---|---|---|---|---|
| 1 |   |   |   |   |
| 2 |   | + |   |   |
| 3 |   |   |   | + |
| 4 |   |   | + | + |

Any type of machine can form a sideways bond with any other type, in both genes and phenes, but a left arm must bond with a right arm (i.e., no left-left nor right-right bonding is allowed). In phenes, the types on each side of a sideways bond govern the angle the bond will take when the strands have folded, as specified in Table 7. This means that the types involved in each bond control the shape that the folded strand will take.

Table 7. Folding angles for sideways bonds between two machine types when in phenes.

| Angle | 1  | 2    | 3   | 4   |
|-------|----|------|-----|-----|
| 1     | 0° | 0°   | 0°  | 0°  |
| 2     | 0° | 120° | 45° | 90° |
| 3     | 0° | 45°  |     |     |
| 4     | 0° | 90°  |     | 60° |

Table 7 shows the sideways bond angle formed by each pair of machine types, when a strand is in its phenotype state; that is, when *folded* is set to 1 (true). When a strand is in its genotype state (*folded* is 0 for all machines in the strand), the sideways bond angles are all 0° (straight). Up bond angles are always 0° ( ignoring random perturbations, from Brownian motion, for example).

The blank cells in Table 7 represent combinations of types for which we have not yet found a use. From Table 6, it can be seen that a type-3 machine cannot form an up bond with another type-3 machine when they are in phenes. Therefore we could control the shape of a 3-3-…-3 phene by specifying any desired value for the angle of 3-3 bonds in Table 7, but the resulting phenes would not be able to form a mesh. Similarly, we could control the shape of a 3-4-3-4-…-3-4 phene by giving any desired value for the angles of 3-4 and 4-3 bonds in Table 7, but the resulting phenes can form up bonds in multiple ways (3-4, 4-3, and 4-4; see Table 6). Thus we have limited control over the shape of the mesh that the phenes will form.

In Table 7, it can be seen that all sideways bonds involving type-1 machines are straight. This allows us to expand the size of a phene, without changing its shape, by inserting a sequence of type-1 machines along each edge. However, it does not allow polygons with exactly two machines on each side, since expansion requires at least one type-1 machine inserted between two other machines.

Given the angles that are available to us in Table 7, some polygons (e.g., octagons and squares) require two types of machines, while others (e.g., triangles and hexagons) involve only one type. We chose to restrict the JohnnyVon 2.0 to four types of machines, in order to demonstrate that a small number of components can be combined to build a variety of structures (like Lego blocks). The angles that we chose make it easy to build triangular meshes. It may seem inconvenient to require two types of machines to build octagonal meshes, but in fact, having two types of machines is helpful with octagons. An octagonal mesh has both octagonal holes and square holes (see Image 4 in Figure 2 in Section 4.2). With two types of machines, we can prevent octagonal phenes from filling in the square holes in the mesh.

A single type of machine would be sufficient to create squares, but using two types permits rectangles that will mesh correctly. With two types of machines, the long and short sides of the rectangle can be distinguished by type, so that two sides will bond together only if they have the same machine type, and thus the same length.

Though they could use two types, hexagons will form a mesh faster with only one type. Furthermore, in order to get the desired behaviour with four machine types, exactly one of the squares or hexagons had to use only one machine type, and the other had to use two. While it would be possible to have an irregular hexagon, this seems much less natural than a rectangle. (In a regular polygon, all sides have the same length. Rectangles are irregular.)

Table 6 shows how machine types control phene bonding in meshes. The bonding rules in Table 6 were designed so that octagonal meshes will form correctly. They prevent octagonal phenes from filling square holes in the mesh.

Additional rules were created to support expansion of the size of a phene, without changing its shape, by inserting a sequence of type-1 machines along each edge. These rules are based on the derived variable *bend-location*. In a phene, a given machine can have either a straight (type 1) or bending (types 2, 3 or 4) machine bonded to its left or right arm. By looking at its neighbours, it can determine where it is in the context of the phene, and hence it can calculate the value of *bend-location*. Certain values of *bend-location* can override the rules in Table 6 and disallow a bond that would otherwise be permitted. Table 8 explains the meaning of the different values of *bend-location*. Table 9 shows how *bend-location* affects bonding.

Table 8. The meaning of the different values of *bend-location*.

| Value | Meaning of value |
|---|---|
| 1 | *right of bend:* right neighbour is straight (1) and left neighbour is bending (2, 3, 4) |
| 2 | *left of bend:* left neighbour is straight (1) and right neighbour is bending (2, 3, 4) |
| 3 | *in bend:* both left and right neighbours are bending types (2, 3, 4) |
| 4 | *extender:* both left and right neighbours are straight types (1) |

Table 9. Pairs of *bend-location* values that will permit an up bond when in phenes.

| bend-location | 1 | 2 | 3 | 4 |
|---|---|---|---|---|
| 1 |   | + |   |   |
| 2 | + |   |   |   |
| 3 |   |   | + |   |
| 4 |   |   |   |   |

When the up field of a machine in one phene overlaps with the up field of a machine in another phene, the rules in both Table 6 and Table 9 must be satisfied before the up fields can bond. If there are no type 1 machines in the phene (i.e., all sideways bonds are bent; the shape is not expanded), then *bend-location* must have the value 3 for all machines, and thus (by Table 9) up bonds depend only on the machine types (Table 6).

In Table 8, we are assuming that the phene forms a closed loop. The error correction system will destroy open loops (Section 3.7.4). When a machine has no left or right neighbour (because it is at the end of an open strand), we treat the missing neighbour as if it were a bending type.

### 3.7.3 Overlap Detection

Because of flexibility in the mesh, and in individual phenes, two phenes can sometimes join a mesh in such a way that their desired positions overlap. This can be most easily seen by imagining a mesh of hexagons that is complete except for a single gap. Suppose that two hexagons jostle (by Brownian motion) into the gap, with slightly different alignments. At roughly the same time, one forms a bond with the phene above the gap, and one with the phene below the gap. As they straighten towards their ideal position (due to twisting forces on their up bonds), each phene may pick up new bonds around the edge of the former gap.

If ignored, this problem spreads, since there are now unbonded up arms on each of the two overlapping hexagons, which permit new hexagons to join the mesh, overlapping those around the former gap, and each of these hexagons in turn can bring in more overlapping hexagons.

To address this problem, an arm was added to the machines (new since JohnnyVon 1.0), called the *overlap detector arm*. In a folded strand, it points towards the center, and will only bond with other overlap detector arms. Both machines must be oriented in the same direction (up to a fixed degree of tolerance), and both machines must have their *in-mesh* flag set to true.

When an overlap detector bond is formed between two machines (necessarily in two different phenes), one machine (chosen arbitrarily) sets its *unfold* signal to true. This signal propagates to its left and right neighbours, setting *folded* to false as it goes. It also breaks the overlap bond that triggered it. The resulting unfolded strand behaves exactly like a newly replicated strand. It tries to replicate until *fold-counter* exceeds its limit, and then it folds up again.

In summary, when two phenes compete for the same gap in a mesh, one of them is forced to become a gene. Converting one of the phenes to a gene, instead of leaving it as a detached phene, allows time for it to drift away from the problematic area, or for the remaining phene to fill up the open bonds in the mesh.

### 3.7.4　Stress Detection

Another type of error can occur in a mesh, again due to the flexibility of the mesh. In this case, we can imagine five triangles bonding to form a pie shape, missing only one more triangle to form a hexagon. However, instead of a new triangle coming in to fill the gap, the two triangles on either side of the gap jostle together, forming a stressed pentagon rather than a hexagon. This pentagonal mesh may be part of a larger mesh, and thus some of the stress may be distributed through the larger mesh. This problem can be partially addressed by increasing the strength of some of the fields and decreasing the tolerance of some of the bonding angles, but it becomes increasingly hard to prevent as the mesh grows.

To detect this kind of problem, each machine maintains a *stress-counter*, which increments each time interval when the machine is not *in-tolerance*, and is reset whenever the machine is *in-tolerance*. This counter can be used to detect cases in which the mesh is stressed because phenes have bonded incorrectly. When the counter exceeds a fixed maximum, it causes the stressed phene to unfold, by setting its *unfold* signal to true. The signal propagates to left and right neighbours, setting *folded* to false and dropping up bonds as it spreads through the neighbours.

### 3.7.5　Seeding the Mesh

In the initial seed gene, *seed-gene* is set to true, but it will be set to false for all of the child genes. If a machine has a true *seed-gene*, it will never trigger the *fold-now* signal. Since the initial seed gene will never fold, there will always be a strand that can continue replicating whenever free machines become available.

The first child of the seed gene, and only the first child of the seed gene, becomes the seed phene. When the *seed-phene* flag is set to true in a strand, it does *not* mean that the given strand is the seed phene; it means that the next child of the given strand will become the seed phene. In the initial seed gene, *seed-phene* is set to true. When the seed gene first replicates, its child examines its parent's *seed-phene* flag and observes that it is set to true. The child then sets it *in-mesh* flag and its *folded* flag to true and it immediately folds to become the seed phene. The parent (the initial seed gene) then sets its *seed-phene* flag to false, so that its future children cannot become seed phenes. (When we say that a strand sets a flag to a value, it is a shorthand way of saying that every machine in the strand sets the flag to the value. Strands do not have flags of their own, other than the flags of their component machines.)

Every other gene, created after the first child, will begin its career with its *in-mesh* flag set to false. If two phenes meet with their *in-mesh* flags set to false, they cannot bond together. A phene can only bond to another phene if the other phene has its *in-mesh* flag set to true. When a machine (in a phene) with a false *in-mesh* flag meets a machine (in a second phene) with a true *in-mesh* flag, they bond (assuming they meet all the conditions in Section 3.7.2), and a signal propagates through the first phene, setting all of the *in-mesh* flags to true (but the signal only propagates from one machine to its sideways neighbour when their bond is *in-tolerance;* see Section 3.7.4). This ensures that the mesh can only grow from the seed phene.

### 3.7.6   Tolerances

Each machine will only form up bonds if all existing bond angles are within a certain tolerance. That is, if a machine's sideways bonds are at angles significantly different from the desired angles (i.e., the angles given by the rules in Section 3.7.2), then no up bonds will form during the current timestep. This prevents unintended up bonds during vulnerable times, such as during splitting or folding.

### 3.7.7   Shattering

There are a number of ways that a gene or phene can break. For example, during splitting, the phase of self-replication when two genes are pushed apart by their repellor arms, if one of them hits the wall of the container at an angle, it puts significant strain on the whole strand. As another example, an error in a mesh can eventually lead to enough strain to pull a phene apart (see Section 3.7.4).

If a machine loses a bond unexpectedly (which is any time other than when splitting or unfolding), or if it notices that its neighbour has folded, then the *shatter* flag is set to true. When a machine observes that its neighbour's *shatter* flag is true, the machine may respond by setting its own *shatter* flag to true. We say that the first machine is the *source* of a *shatter* signal that was *received* by the second machine.

The *shatter* signal always propagates through sideways bonds, setting the *shatter* flag to true in *left-neighbours* and *right-neighbours*. The shatter signal may also propagate to an *up-neighbour*, but only if the source machine has *replicated* but not *folded*. If the neighbouring machine has *replicated*, something went wrong with the split (two *replicated* machines should not be bonded before they're both folded); on the other hand, if the neighbour's *replicated* flag is false, then it may be part of an incomplete copy, and thus should be abandoned.

When a machine's *shatter* flag is true, it drops all of its bonds (the discrete timesteps ensure that the state is propagated, even if the bonds are broken during that time step, since machines consult their neighbours' state as it was at the beginning of the step). When the bonds have been dropped, it then sets *folded, seed-gene,* and *replicated* flags to false and becomes a free machine.

The *shatter* mechanism is not a subtle way to handle errors, but we have found it to be effective. In our simulations, shattering is relatively rare. This error correction mechanism is similar to Sayama's method for handling errors in self-replicating loops [12], [13].

## 3.8　Implementation

JohnnyVon 2.0 builds directly on the original JohnnyVon 1.0. Both systems are written in Java and their source code is available under the GNU General Public License (GPL) at http://purl.org/net/johnnyvon/.

# 4　Experiments and Discussion

In our first experiment, we demonstrate the construction of a small mesh of triangles, highlighting several important points in the replication and assembly. In the next set of experiments, we demonstrate replication and assembly of meshes built from each of the supported polygons, with one machine per side. We then show a mesh of polygons with more than one machine per side, a 3×1 rectangle and a triangle with three machines per side. Finally, to demonstrate scalability, we show a large mesh of triangles.

In the following figures, the inner grey square represents the container. The middle of a machine must stay inside the grey square. (It takes less computation to check whether the middles are within bounds than to check all of the arms.)

## 4.1　Self-Replication and Self-Assembly

In Figure 1, the images show a typical run of JohnnyVon 2.0. The run starts with a soup of 54 free type-2 machines and a seed gene of the form 2-2-2, and it ends with a triangular mesh.

**Image 1:** This shows the initial configuration. Each of the free machines is in a random position and the seed gene is in the center (it is the strand of three machines, forming a straight line).

**Image 2:** After 2,385 steps, the first replication is complete. We see two genes, immediately after they have split and their repellor arms have pushed them apart.

**Image 3**: The first child of the seed gene is folding up, to become the seed phene. The seed gene has already begun a second copy.

**Image 4:** By time 44,235, nearly all of the free machines are now attached to genes. Since there are so few free machines left, most of these genes cannot complete self-replication. As some of the incomplete strands' *fold-counters* hit their upper limit, they will fold and release free machines, allowing other genes to complete self-replication.

**Image 5:** We can see the second phene forming. In this image, it has not completely folded; the triangle has a small gap at the top.

**Image 6:** Slightly more than 3,000 timesteps later, the new phene has bonded with the seed phene.

**Image 7:** Now many more triangles have folded and joined the mesh. Two triangles have not yet joined (one is in the lower left corner and the other is near the center).

**Image 8:** The mesh is almost complete. In the bottom on the right, there is a pentagonal arrangement of five triangles. This would eventually be corrected (by one of the triangles releasing and unfolding; see Section 3.7.4), although the container is just barely large enough to hold a mesh that includes all of the machines, and thus errors may continue to form even as they are corrected. In a situation where the container constrains the mesh, it is possible for a machine to get attached to a mesh in such a way that it can never reach an equilibrium where all of its bonds are in tolerance, since the conditions for accepting new bonds are much looser than the conditions for detecting stress.

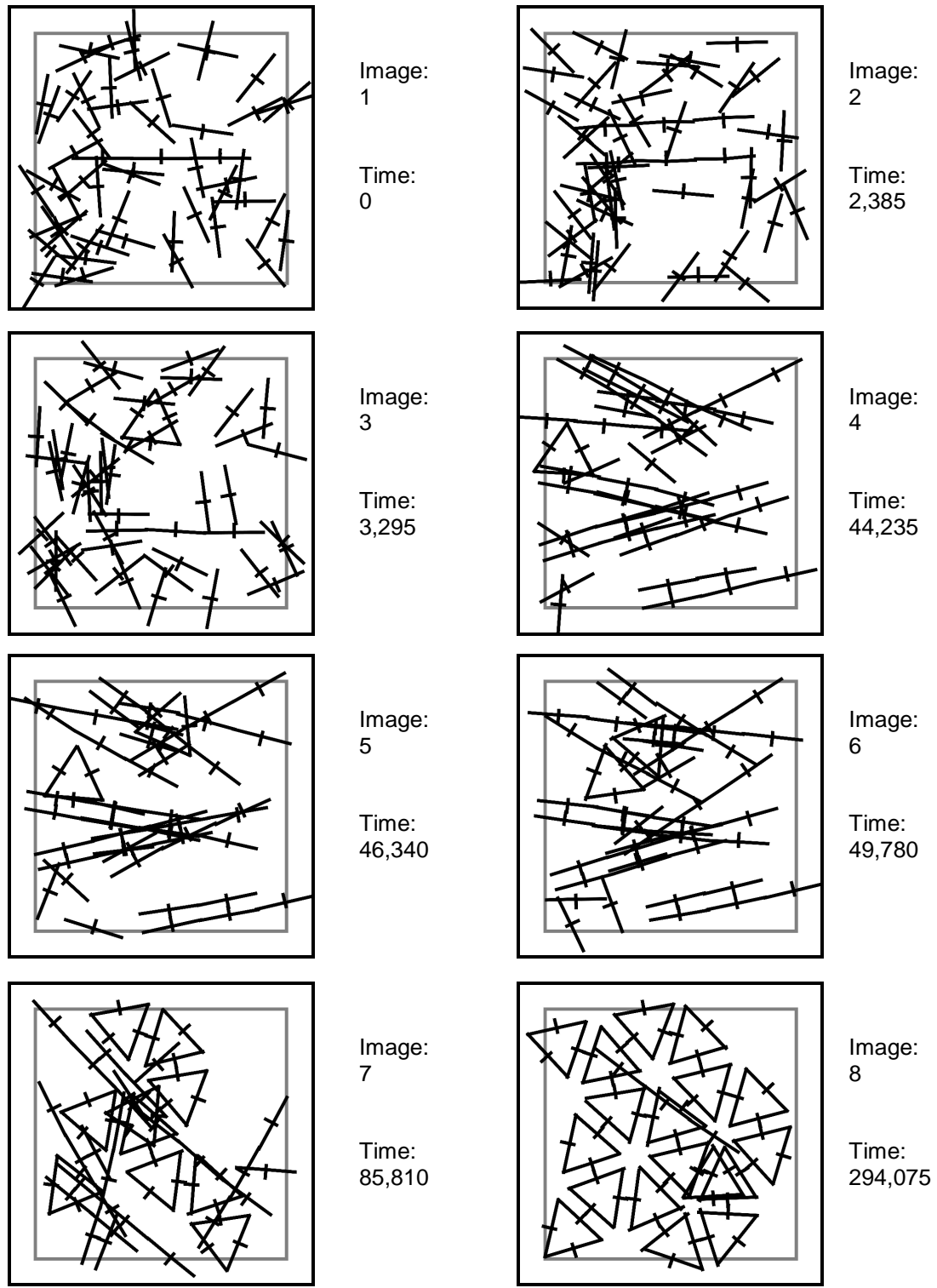

Figure 1. These images illustrate the experiment described in Section 4.1.

In this simulation, the container is relatively small, and therefore Brownian motion is relatively strong. With strong Brownian motion, free machines are quickly distributed throughout the container, thus a replicating strand has a steady supply of free machines. The small container also means that the phenes never have far to go to join the mesh, and will quickly be bumped into the right position. In a larger container, replicating strands will consume the machines in their local area, and then replication slows until diffusion replenishes the supply. It also takes longer for phenes to find a place where they can join the mesh. We could speed up the action in a larger container by increasing the Brownian motion (i.e., turning up the heat), but that could damage the mesh.

### 4.2 Simple Polygonal Meshes

In Figure 2, we show assembled mesh structures. Four different regular polygonal meshes are shown, all with sides that are one machine in length. Each of these four simulations was started with a single seed strand and was executed until the mesh was well developed. The scale of the images in Figure 2 is different from the scale of the images in Figure 1. These simulations use a container about nine times larger in area than the simulations in Figure 1. Table 10 summarizes the four simulations in Figure 2.

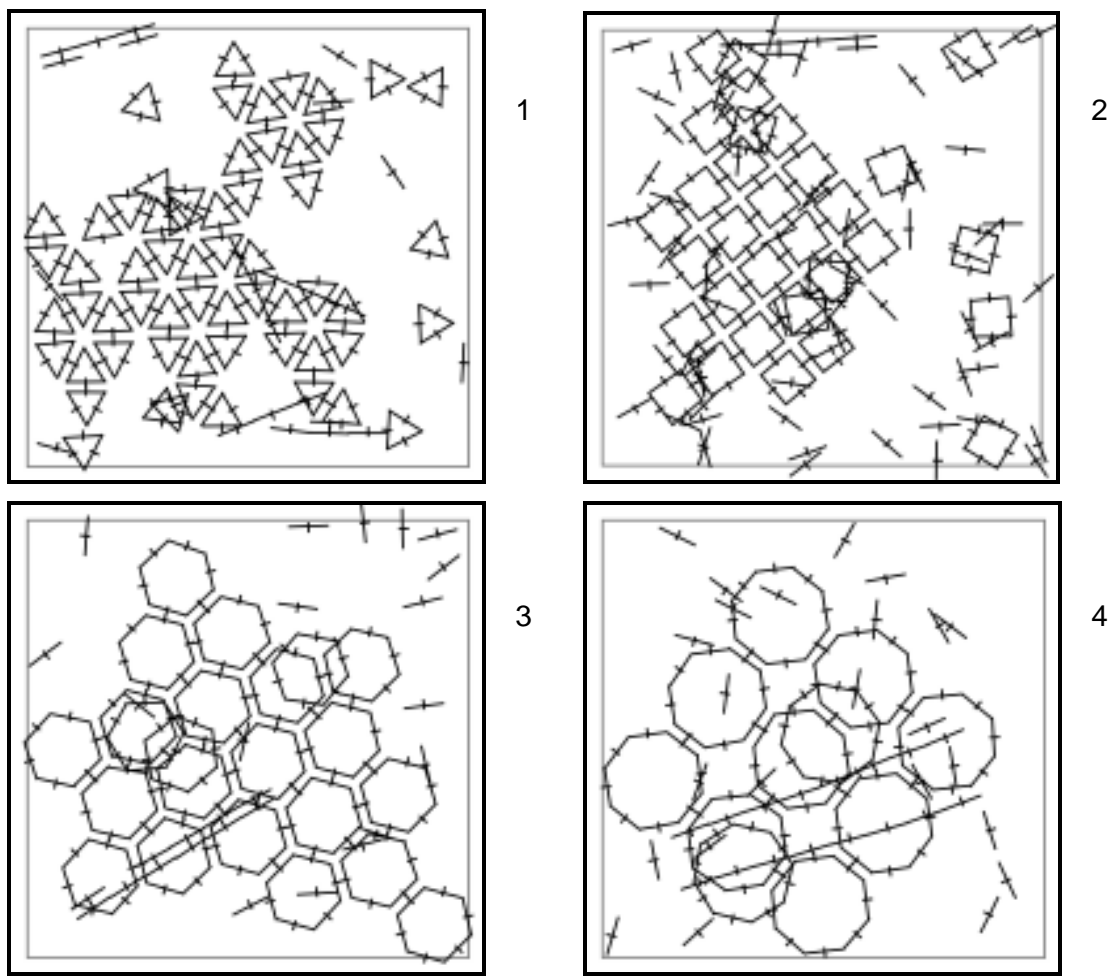

Figure 2. The resulting mesh for each of the four supported regular polygons

Table 10. Some basic observations about each image in Figure 2.

| Image | Phenes | Seed gene | Timestep | Initial free machines | Phenes in mesh | Genes remaining |
|---|---|---|---|---|---|---|
| 1 | Triangles | 2-2-2 | 246,000 | 201 | 50 | 5 |
| 2 | Squares | 4-2-4-2 | 498,600 | 200 | 24 | 1 |
| 3 | Hexagons | 4-4-4-4-4-4 | 448,600 | 160 | 20 | 1 |
| 4 | Octagons | 2-3-2-3-2-3-2-3 | 1,107,400 | 120 | 9 | 2 |

## 4.3 Fancy Meshes

In Figure 3, Image 1 (timestep 691,900) shows a mesh built of rectangles, rather than regular polygons. Because squares and rectangles use two types of machines (see Section 3.7.2), the rectangles only join the mesh if they are correctly oriented. The seed for Image 1 was 2-4-2-1-2-4-2-1.

Image 2 (timestep 78,800) shows large triangles. The seed was 2-1-2-2-1-2-2-1-2. The bonds between type-2 machines fold to form the corners, while the type-1 machines provide the extension to make these triangles larger. In principle, each phene can be made arbitrarily large using this approach.

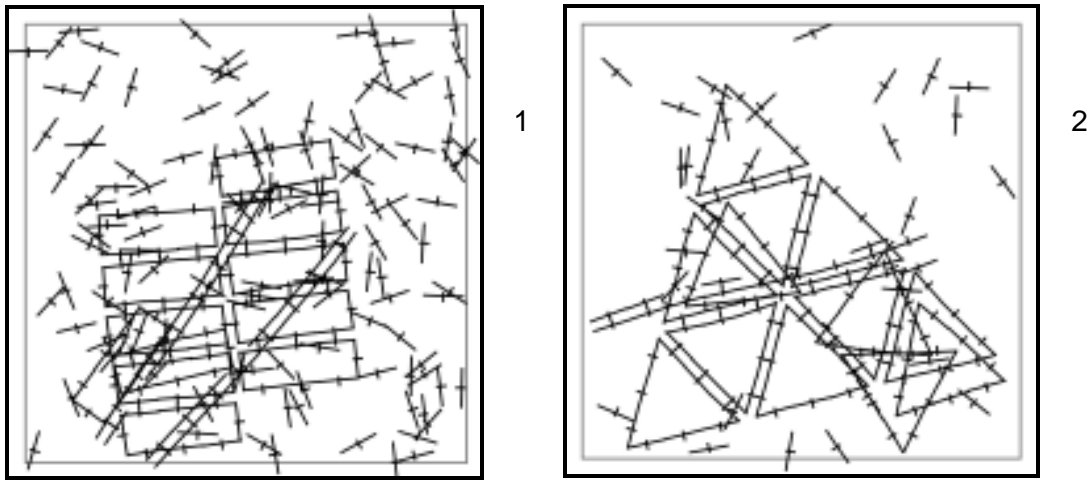

Figure 3. A rectangular mesh and a mesh of expanded triangles.

## 4.4 Large Mesh

The image in Figure 4 demonstrates that meshes can grow correctly beyond a small number of triangles. The seed was 2-2-2. The mesh contains 234 triangles.

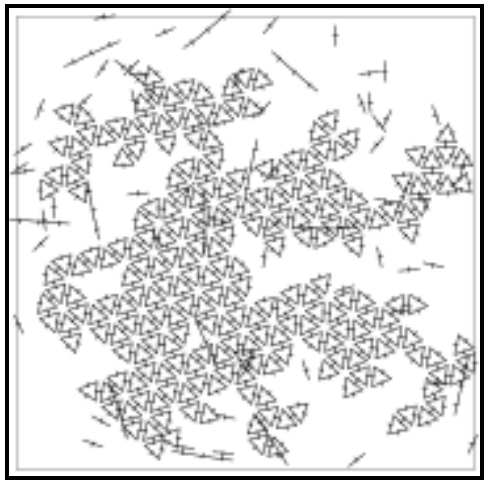

Figure 4. A mesh of 234 triangles.

## 5  Limitations and Future Work

JohnnyVon 2.0 has several minor limitations. For example, phenes must be closed for the system to work correctly. Although a hexagon composed of five machines and a gap in the sixth side can form a mesh, the error correction system would destroy the resulting mesh. Because closure of the phenes increases their rigidity, a mesh built of open phenes would be more flexible, and there may be other interesting effects.

The variety of phenes in JohnnyVon 2.0 is also somewhat limited. Our original goal, to support all regular polygons that tile the plane (triangles, squares and hexagons), is satisfied. JohnnyVon 2.0 also supports partial tiling with octagons (square gaps are left in the mesh), and full tiling with rectangles. However, we would now like to support concave shapes (e.g., stars), as well as more general polygons. It would be interesting to enable Penrose tilings and Kepler tilings [3].

The replication phase takes much longer with two (or more) types of machines than it does with one, since each free machine has fewer places to bond correctly (equivalently, each machine in the strand has fewer free machines available with which it can bond). Supplying two (or more) times as many machines increases the computation per timestep (roughly quadratically). However, JohnnyVon 2.0 should be parallelizable. This is another area for future work.

Like JohnnyVon 1.0, version 2.0 still runs on a standard desktop computer, thanks in part to improvements in hardware since the development of version 1.0. However, there were many experiments we wanted to try (e.g., polygons with 4 or 5 machines per side) that were not practical, given our available hardware and our patience. This problem can be addressed by improving the efficiency of our implementation, converting the code to a more efficient language than Java (which is likely to make it much less portable), parallelizing the code, or obtaining better hardware.

The computational complexity of the simulation increases with the size of the phenes, since each phene must be jostled to a place near where it belongs, and larger phenes move more slowly. Meshes of large phenes require many timesteps to be constructed. The problem may be alleviated by increasing the Brownian motion or decreasing the viscosity of the simulated liquid, but each of these solutions presents new problems. We have tuned the physical parameters, in an effort to balance these conflicting concerns.

The current settings of the physical parameters appear to strike a good balance, but there is likely room for further improvements.

Referring to Table 1 in Section 2.4, JohnnyVon would benefit from increased realism and increased programmability. Although JohnnyVon 2.0 provides a moderate level of programmability, it is not as programmable as we would like. One problem is that the mesh grows without control. Sometimes the mesh is relatively dense (as in Image 2 of Figure 2) while at other times it has many gaps (as in Figure 4). We would like to add a programmable mechanism for controlling the final size and shape of the mesh, and for avoiding meshes with large gaps (or for deliberately creating gaps, which may be useful for some applications).

In the context of JohnnyVon's virtual physics, it may be meaningful to define a universal constructor. For example, we might say that a universal constructor would be capable of building any two-dimensional structure that can be constructed from a finite number of machines, such that up arms are bonded to up arms and left arms are bonded to right arms. The design for the structure should be encoded in a seed gene. Ideally, the seed would contain many fewer machines than the final structure, although this may not be possible when the final structure lacks a regular pattern. Much further work is required to make a universal constructor in the JohnnyVon model.

JohnnyVon 2.0 also provides a moderate level of realism, but again it is not as realistic as we would like. Our attractive and repulsive forces are somewhat unlike electrical or magnetic attraction and repulsion. The JohnnyVon simulation also does not attempt to model conservation of energy. Arbesman has recently done some interesting work on computational simulation of artificial life with conservation of energy [1].

Other steps towards increased realism would be to extend the simulation to three dimensions and to model the physics of the internal operations of the machines. Currently the external relations between machines are governed by a simple virtual physics, but the internal operations are described by abstract finite automata. However, both of these steps to realism would involve a significant increase in computational complexity.

## 6  Applications

With JohnnyVon 2.0, we have focused more clearly on nanotechnology, at the expense of application to theoretical biology. In our previous work, we suggested that JohnnyVon 1.0 provided a plausible mechanism for nanoscale manufacturing [18]. A vat of liquid containing free machines would be seeded with a single strand, soon resulting in a vat full of copies of the seed strand. JohnnyVon 2.0 takes this application one step further, beyond self-replication to programmable construction of meshes. Since the user has some control over the size and shape of the holes in the mesh, we can imagine these meshes being produced for filtration, insulation, or simply as kind of cloth.

If we can create a mechanism for controlling the size and shape of the mesh, more applications become possible. Since the system is accurate and self-correcting, pieces of cloth could be created exactly to specification, down to the size of a single machine.

## 7  Conclusion

JohnnyVon 1.0 demonstrated self-replication in a continuous two-dimensional space with virtual physics. JohnnyVon 2.0 goes beyond its predecessor by introducing a user-programmable phenotype, consisting of a variety of meshes. JohnnyVon 2.0 is more realistic than cellular automata models [6], [11], [12], [13], [19], more programmable than artificial chemistry models [4], [5], and more computationally tractable than von Neumann's universal constructor [10], [20]. However, there is still much room for improvement in the degree of physical realism of the simulation and in the degree of programmability of the phenotype.

Like its predecessor, JohnnyVon 2.0 is a local model. There is no global data structure that represents strands or meshes; these are emergent entities that arise from the interactions of the basic elements (the machines). Each machine is autonomous and can only sense its immediate neighbours. Control is local, distributed, and parallel.

From four different types of machines, JohnnyVon can produce four different polygonal meshes, with an infinite number of possible sizes (as per Section 4.3). The user can specify the mesh that will be produced by encoding the desired size and shape in the initial seed, without making any changes to the physics of the simulation. Errors in replication and in mesh formation are automatically detected and corrected, using purely local mechanisms.

JohnnyVon 2.0 also avoids the "grey goo" scenario of self-replicating nanobots run amok. Replication and assembly are inherently limited by the supply of machines; when the free machines have all bonded, the process stops.

## Acknowledgements

Thanks to Arnold Smith for starting us down this path, with JohnnyVon 1.0.